\documentclass{article}

\usepackage{graphicx}

\usepackage[round,colon]{natbib}

\usepackage{color}
\usepackage{amsmath}
\usepackage{amssymb}
\usepackage{amsfonts}
\usepackage{epsfig}
\usepackage{url}
\usepackage{bm}

\def\bfe{\mbox{\boldmath $e$}}
\def\bff{\mbox{\boldmath $f$}}
\def\bfx{\mbox{\boldmath $x$}}
\def\bfy{\mbox{\boldmath $y$}}
\def\bfz{\mbox{\boldmath $z$}}

\def\bfu{\mbox{\boldmath $u$}}

\def\bfmu{\mbox{\boldmath $\mu$}}

\def\bSigma{\mbox{\boldmath $\Sigma$}}
\def\btheta{\mbox{\boldmath $\theta$}}
\def\beta{\mbox{\boldmath $\eta$}}

\newcommand{\B}{{\bf B}}
\newcommand{\C}{{\bf C}}

\newcommand{\bLambda}{{\bf \Lambda}}
\begin{document}

\title{Bayesian estimation of possible causal
direction in the presence of latent confounders using a linear
non-Gaussian acyclic structural equation model with
individual-specific effects}
\author{Shohei Shimizu\thanks{The Institute of Scientific and Industrial Research (ISIR), Osaka University,
Mihogaoka 8-1, Ibaraki, Osaka 567-0047, Japan. Email: sshimizu@ar.sanken.osaka-u.ac.jp}, Kenneth Bollen\thanks{Department of Sociology, CB 3210 Hamilton Hall, University of North Carolina, Chapel Hill, NC 27599-3210, U.S.A. Email: bollen@unc.edu}}
\date{}

\maketitle

\begin{abstract}
Several existing methods have been shown to consistently estimate causal direction assuming linear or some form of nonlinear relationship and no latent confounders. 
However, the estimation results could be distorted if either assumption is violated.
{\color{black} We develop an approach to determining the possible causal direction between two observed variables when latent confounding variables are present.} 
We first propose a new linear non-Gaussian acyclic structural equation model with individual-specific effects {\color{black} {\color{black} that are sometimes} the source of confounding. 
Thus, modeling individual-specific effects as latent variables allows latent confounding to be considered. }
We then propose an empirical Bayesian approach for estimating possible causal direction using the new model.
We demonstrate the effectiveness of our method using artificial and real-world data.
\end{abstract}

\section{Introduction}\label{sec:intro}
{\color{black} Aids to uncover the causal structure of variables from observational data are welcomed additions to the field of machine learning \citep{Pearl00book,Spirtes93book}.}
One conventional approach {\color{black} makes use of} Bayesian networks \citep{Pearl00book,Spirtes93book}.
However, these suffer from the identifiability problem. {\color{black} That is}, many different causal structures give the same conditional independence between variables, and in many cases one cannot uniquely estimate the underlying causal structure without prior knowledge~\citep{Pearl00book,Spirtes93book}.

{\color{black} To address these issues,} \citet{Shimizu06JMLR} proposed LiNGAM (Linear Non-Gaussian Acyclic Model), a variant of Bayesian networks \citep{Pearl00book,Spirtes93book} and structural equation models \citep{Bollen89book}.
Unlike conventional Bayesian networks, LiNGAM is a fully identifiable model \citep{Shimizu06JMLR}, and has recently attracted much attention in machine learning \citep{Spirtes10Tribute,Moneta11NIPS09}.
{\color{black} If causal relations exist among variables, LiNGAM uses their non-Gaussian distributions to identify the causal structure among the variables. }
LiNGAM is closely related to independent component analysis (ICA) \citep{Hyva01book}; the identifiability proof and estimation algorithm are partly based on the ICA theory. 
The idea of LiNGAM has been extended in many directions, including to nonlinear cases \citep{Hoyer09NIPS,Lacerda08UAI,Hyva10JMLR,Zhang09UAI,Peters11TPAMI}.

Many causal discovery methods including LiNGAM make the strong assumption of no latent confounders \citep{Spirtes91,Dodge01AmericanStatistician,Shimizu06JMLR,Hyva13JMLR,Hoyer09NIPS,Zhang09UAI}.
These methods have been used in various application fields \citep{Ramsey13NeuroImage,Rosenstrom12PlosOne,Smith11NeuroImage,Statnikov12BMC,Moneta12OxfordBulletin}.
However, in many areas of empirical science, it is often difficult to accept the estimation results because latent confounders are ignored.
In theory, we could take a non-Gaussian approach \citep{Hoyer08IJAR} that uses an extension of ICA with more latent variables than observed variables (overcomplete ICA) to formally consider latent confounders in the framework of LiNGAM. Unfortunately, current versions of the overcomplete ICA algorithms are not very computationally reliable since they often suffer from local optima \citep{Entner10AMBN}.

Thus, in this paper, we propose an alternative Bayesian approach to develop a method that is computationally simple in the sense that no iterative search in the parameter space is required
and it is capable of finding the possible causal direction of two observed variables in the presence of latent confounders.
We first propose a variant of LiNGAM with individual-specific effects.
Individual differences are {\color{black} sometimes} the source of confounding {\color{black}\citep{vonEye03DP}}. 
Thus, modeling certain individual-specific effects as latent variables allows a type of latent confounding to be considered. 
A latent confounding variable is an unobserved variable that exerts a causal influence on more than one observed variables \citep{Hoyer08IJAR}.
The new model is still linear but allows any number of latent confounders. 
We then present a Bayesian approach for estimating the model by integrating out some of the large number of parameters, which is of the same order as the sample size.
Such a Bayesian approach is often used in the field of mixed models \citep{Demidenko04book} {\color{black} and multilevel models \citep{Kreft98book}}, although estimation of causal direction is not a topic studied within it. 

{\color{black} Granger causality \citep{Granger69} is another popular method to aid detection of causal direction. His method depends on the temporal ordering of variables whereas our method does not. Therefore, our method can be applied to cases where temporal information is not available, i.e., cross-sectional data, as well as those where it is available, i.e., time-series data.}

The remainder of this paper is organized as follows.
We first review LiNGAM \citep{Shimizu06JMLR} and its extension to latent confounder cases \citep{Hoyer08IJAR} in Section~\ref{sec:previous}.
In Section~\ref{sec:proposal}, we propose a new mixed-LiNGAM model, which is a variant of LiNGAM with individual-specific effects. We also propose an empirical Bayesian approach for learning the model.
We empirically evaluate the performance of our method using artificial and real-world sociology data in Sections~\ref{sec:exp} and~\ref{sec:real}, respectively, and present our conclusions in Section~\ref{sec:conc}.

\section{Background}\label{sec:previous}
In this section, we first review the linear non-Gaussian structural equation model known as LiNGAM \citep{Shimizu06JMLR}.
We then discuss an extension of LiNGAM to cases where latent confounding variables exist \citep{Hoyer08IJAR}.

In LiNGAM \citep{Shimizu06JMLR}, causal relations between observed variables $x_l$ ($l=1,\cdots,d$) are modeled as
\begin{eqnarray}
x_l &=& \mu_l + \sum_{k(m)<k(l)} b_{lm}x_m + e_l, \label{eq:model1}
\end{eqnarray}
where $k(l)$ is a causal ordering of the variables $x_l$. 
{\color{black} The causal orders $k(l)$ ($l=1, \cdots, d$) are unknown and to be estimated.}
In this ordering, the variables $x_l$ form a directed acyclic graph (DAG) so that no later variable determines, i.e., has a directed path to, any earlier variable in the DAG.
The variables $e_l$ are latent continuous variables called error variables, $\mu_l$ are intercepts or regression constants, and $b_{lm}$ are connection strengths or regression coefficients. 

\begin{figure}[tb]
\begin{center}
\includegraphics[width=0.2\textwidth]{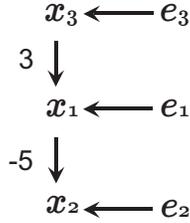}
\caption{An example graph of LiNGAMs}
\label{fig:expsem}
\end{center}
\end{figure}

In matrix form, the LiNGAM model in Eq.~(\ref{eq:model1}) is written as
\begin{eqnarray}
\bfx &=& \bfmu+\B\bfx + \bfe,\label{eq:model2}
\end{eqnarray}
where the vector $\bfmu$ collects constants $\mu_l$, the connection strength matrix $\B$ collects regression coefficients (or connection strengths) $b_{lm}$, and the vectors $\bfx$ and $\bfe$ collect observed variables $x_l$ and error variables $e_l$, respectively.
The zero/non-zero pattern of $b_{lm}$ corresponds to the absence/existence pattern of directed edges (direct effects). 
It can be shown that it is always possible to perform simultaneous, equal row and column permutations on the connection strength matrix $\B$ to cause it to become {\it strictly} lower triangular, based on the acyclicity assumption \citep{Bollen89book}. Here, strict lower triangularity is defined as a lower triangular structure with the diagonal consisting entirely of zeros. 
{\color{black} Errors $e_l$ follow non-Gaussian distributions with zero mean and non-zero variance, and are {\color{black} jointly independent}. }
This model {\color{black} without assuming non-Gaussianity distribution} is called a fully recursive model in conventional structural equation models \citep{Bollen89book}.
The non-Gaussianity assumption on $e_l$ enables the identification of a causal ordering $k(l)$ {\color{black} and the coefficients $b_{lm}$} 
based only on $\bfx$~\citep{Shimizu06JMLR}, unlike conventional Bayesian networks based on the Gaussianity assumption on $e_l$ \citep{Spirtes93book}. 
To illustrate the LiNGAM model, the following example is considered, whose corresponding directed acyclic graph is provided in Fig.~\ref{fig:expsem}: 
\begin{eqnarray}
\left[
\begin{array}{c}
x_1\\
x_2\\
x_3
\end{array}
\right]
&=&
\left[
\begin{array}{ccc}
0 & 0 & 3\\
-5 & 0 & 0\\
0 & 0 & 0
\end{array}
\right]
\left[
\begin{array}{c}
x_1\\
x_2\\
x_3
\end{array}
\right]
+
\left[
\begin{array}{c}
e_1\\
e_2\\
e_3
\end{array}
\right].
\end{eqnarray}
In this example, $x_3$ is {\it equal to} error $e_3$ and is exogenous since it is not affected by either of the other two variables $x_1$ and $x_2$. 
Thus, $x_3$ is in the first position of such a causal ordering such that $\B$ is strictly lower triangular, $x_1$ is in the second, and $x_2$ is the third, i.e., $k(3)=1$, $k(1)=2$, and $k(2)=3$. 
If we permute the variables $x_1$ to $x_3$ according to the causal ordering, we have
\begin{eqnarray}
\left[
\begin{array}{c}
x_3\\
x_1\\
x_2
\end{array}
\right]
&=&
\left[
\begin{array}{ccc}
0 & 0 & 0\\
3 & 0 & 0\\
0 & -5 & 0
\end{array}
\right]
\left[
\begin{array}{c}
x_3\\
x_1\\
x_2
\end{array}
\right]
+
\left[
\begin{array}{c}
e_3\\
e_1\\
e_2
\end{array}
\right].
\end{eqnarray}
It can be seen that the resulting connection strength {\color{black} (or regression coefficient)} matrix is strictly lower triangular. 

Several computationally efficient algorithms for estimating the model have been proposed \citep{Shimizu06JMLR,Shimizu11JMLR,Hyva13JMLR}.
{\color{black}As with ICA, LiNGAM is identifiable under the assumptions of non-Gaussianity and independence among error variables \citep{Shimizu06JMLR,Comon94SP,Eriksson03ICA}}.\footnote{\citet{Comon94SP} and \citet{Eriksson03ICA} established the identifiability of ICA based on the characteristic functions of variables. Moments of some variables may not exist, but their characteristic functions always exist.}
However, for the estimation methods to be consistent, additional assumptions, e.g., the existence of their moments or some other statistic, must be made to ensure that the statistics computed in the estimation algorithms exist. 
The idea of LiNGAM can be generalized to nonlinear cases \citep{Hoyer09NIPS,Tillman09NIPS,Zhang09UAI,Peters11UAI}.

The assumption of independence among $e_l$ means that there is no latent confounding variable \citep{Shimizu06JMLR}.
A latent confounding variable is an unobserved variable that contributes to the values of more than one observed variable \citep{Hoyer08IJAR}. 
However, in many applications, there often exist latent confounding variables.
If such latent confounders are completely ignored, the estimation results can be seriously biased \citep{Pearl00book,Spirtes93book,Bollen89book}.
Therefore, in \citet{Hoyer08IJAR}, LiNGAM with latent confounders, called latent~variable~LiNGAM, was proposed, and the model can be formulated as follows:
\begin{eqnarray}
x_l &=& \mu_l + \sum_{k(m)<k(l)} b_{lm}x_m + \sum_{q=1}^Q \lambda_{lq}f_q + e_l, \label{eq:lvlingam0}
\end{eqnarray}
where $f_q$ are non-Gaussian individual-specific effects $f_q$ with zero mean and unit variance and $\lambda_{lq}$ denote the regression coefficients (connection strengths) from $f_q$ to $x_l$. 
This model is written in matrix form as follows:
\begin{eqnarray}
\bfx &=& \bfmu + \B\bfx + \bLambda \bff + \bfe,\label{eq:lvlingam}
\end{eqnarray}
where the difference from LiNGAM in Eq.~(\ref{eq:model2}) is the existence of a latent confounding variable vector $\bff$.
The vector $\bff$ collects $f_q$. 
The matrix $\bLambda$ collects $\lambda_{lq}$ and is assumed to be of full column rank. 
{\color{black}
Another way to represent latent confounder cases would be to use dependent error variables. Denoting $\bLambda \bff+\bfe$ in Eq.~(\ref{eq:lvlingam}) by $\tilde{\bfe}$, we have
\begin{eqnarray}
\bfx &=& \bfmu + \B\bfx + \bLambda \bff + \bfe \\
&=& \bfmu + \B\bfx + \tilde{\bfe},
\end{eqnarray}
where $\tilde{e}_i$ are dependent due to the latent confounders $f_q$. 
Observed variables that are equal to dependent errors $\tilde{e}_i$ are connected by bi-directed arcs in their graphs. An example graph is given in Fig.~\ref{fig:domain}. 
This representation can be more general since it is easier to extend it to represent nonlinearly dependent errors. 
In this paper, however, we use the aforementioned representation using independent errors and latent confounders since linear relations of the observed variables, latent confounders, and errors are necessary for our approach. 
}

\begin{figure}[t]
\begin{center}
\includegraphics[width=.75\textwidth]{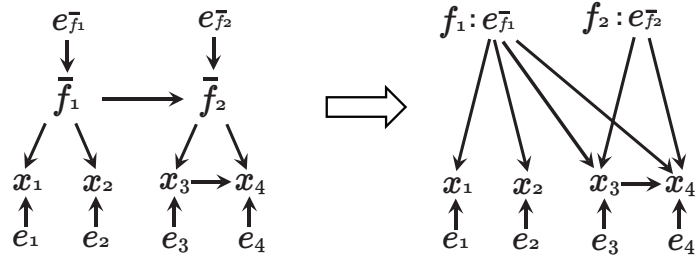}
\caption{An example graph to illustrate the idea of independent latent confounders.}
\label{fig:example1}
\end{center}
\end{figure}

Without loss of generality, the latent confounders $f_q$ are assumed to be {\color{black} jointly independent} since any dependent latent confounders can be remodeled by linear combinations of independent latent variables if the underlying model is linear acyclic and the error variables are independent \citep{Hoyer08IJAR}. 
To illustrate this, the following example model is considered:
\begin{eqnarray}
\bar{f}_1 &=& e_{\bar{f}_1} \label{eq:exp1_1_0}\\
\bar{f}_2 &=& \omega_{21}\bar{f}_1 + e_{\bar{f}_2} \label{eq:exp1_1_1}\\
x_1 &=& \lambda_{11}\bar{f}_1 + e_1 \label{eq:exp1_1_2}\\
x_2 &=& \lambda_{21}\bar{f}_1 + e_2 \label{eq:exp1_1_3}\\
x_3 &=& \lambda_{32}\bar{f}_2+ e_3 \label{eq:exp1_1_4} \\
x_4 &=& b_{43}x_3 + \lambda_{42}\bar{f}_2+ e_4, \label{eq:exp1_1_5}
\end{eqnarray}
where errors $e_{\bar{f}_1}$(=$\bar{f}_1$), $e_{\bar{f}_2}$, and $e_1$--$e_4$ are non-Gaussian and independent. The associated graph is shown in Fig.~\ref{fig:example1}. The relations of $\bar{f}_1$, $\bar{f}_2$, and $x_1$--$x_4$ are represented by a directed acyclic graph and latent confounders $\bar{f}_1$ and $\bar{f}_2$ are dependent. 
In matrix form, this example model can be written as
\begin{eqnarray}
\left[
\begin{array}{c}
x_1 \\
x_2 \\
x_3 \\
x_4 
\end{array}
\right]
=
\left[
\begin{array}{cccc}
0 & 0 & 0 & 0 \\
0 & 0 & 0 & 0 \\
0 & 0 & 0 & 0 \\
b_{43} & 0 & 0 & 0
\end{array}
\right]
\left[
\begin{array}{c}
x_1 \\
x_2 \\
x_3 \\
x_4 
\end{array}
\right]
+
\left[
\begin{array}{cc}
\lambda_{11} & 0 \\
\lambda_{21} & 0 \\
0 & \lambda_{32} \\
0 & \lambda_{42} \\
\end{array}
\right]
\left[
\begin{array}{c}
\bar{f}_1 \\
\bar{f}_2 
\end{array}
\right]
+
\left[
\begin{array}{c}
e_1 \\
e_2 \\
e_3 \\
e_4 
\end{array}
\right].
\end{eqnarray}
The relations of $\bar{f}_1$ and $\bar{f_2}$ to $e_{\bar{f}_1}$ and $e_{\bar{f}_2}$ in Eqs.~(\ref{eq:exp1_1_0})--(\ref{eq:exp1_1_1}):
\begin{eqnarray}
\left[
\begin{array}{c}
\bar{f}_1 \\
\bar{f}_2 
\end{array}
\right]
=
\left[
\begin{array}{cc}
1 & 0 \\
\omega_{21} & 1
\end{array}
\right]
\left[
\begin{array}{c}
e_{\bar{f}_1} \\
e_{\bar{f}_2} 
\end{array}
\right],
\end{eqnarray}
we obtain
\begin{eqnarray}
\underbrace{
\left[
\begin{array}{c}
x_1 \\
x_2 \\
x_3 \\
x_4 
\end{array}
\right]
}_{\bfx}
&=&
\underbrace{
\left[
\begin{array}{cccc}
0 & 0 & 0 & 0 \\
0 & 0 & 0 & 0 \\
0 & 0 & 0 & 0 \\
b_{43} & 0 & 0 & 0
\end{array}
\right]
}_{\B}
\underbrace{
\left[
\begin{array}{c}
x_1 \\
x_2 \\
x_3 \\
x_4 
\end{array}
\right]
}_{\bfx}
\nonumber \\
& & \hspace{5mm} 
+
\underbrace{
\left[
\begin{array}{cc}
\lambda_{11} & 0 \\
\lambda_{21} & 0 \\
\lambda_{32}\omega_{21} & \lambda_{32} \\
\lambda_{42}\omega_{21} & \lambda_{42} \\
\end{array}
\right]
}_{\bLambda}
\underbrace{
\left[
\begin{array}{c}
e_{\bar{f}_1} \\
e_{\bar{f}_2} 
\end{array}
\right]
}_{\bff}
+
\underbrace{
\left[
\begin{array}{c}
e_1 \\
e_2 \\
e_3 \\
e_4 
\end{array}
\right]
}_{\bfe}.
\end{eqnarray}
This is a latent~variable~LiNGAM in Eq.~(\ref{eq:lvlingam}) taking $f_1=e_{\bar{f}_1}$ and $f_2=e_{\bar{f}_2}$ since $e_{\bar{f}_1}$ and $e_{\bar{f}_2}$ are non-Gaussian and independent.

Moreover, the faithfulness of $x_l$ and $f_q$ to the generating graph is assumed. 
The faithfulness assumption \citep{Spirtes93book} here means that when multiple causal paths exist from one variable to another, their combined effect does not equal exactly zero \citep{Hoyer08IJAR}. 
The faithfulness assumption can be considered to be not very restrictive from the Bayesian viewpoint \citep{Spirtes93book} since the probability of having exactly the parameter values that do not satisfy faithfulness is zero \citep{Meek95UAI}.

In the framework of latent~variable~LiNGAM, it has been shown \citep{Hoyer08IJAR} that the following three models are distinguishable based on observed data\footnote{If one or more error variables or latent confounders are Gaussian, it cannot be ensured that Models~3 to 5 will be distinguishable. \citet{Hoyer08UAI} considered cases with one or more Gaussian error variables in the context of basic LiNGAM.}, i.e., the three different causal structures induce different data distributions:
\begin{eqnarray}
& &{\rm Model}~3: \left\{
\begin{array}{l}
x_1 = \hspace{12.5mm} \sum_{q=1}^Q \lambda_{1q}f_q + e_1\\
x_2 = \hspace{12.5mm} \sum_{q=1}^Q \lambda_{2q}f_q + e_2,
\end{array}
\right.\\
& &{\rm Model}~4: \left\{
\begin{array}{l}
x_1 = \hspace{12.5mm} \sum_{q=1}^Q \lambda_{1q}f_q + e_1\\
x_2 = b_{21} x_1 + \sum_{q=1}^Q \lambda_{2q}f_q + e_2,
\end{array}
\right.\\
& &{\rm Model}~5: \left\{
\begin{array}{l}
x_1 = b_{12} x_2 + \sum_{q=1}^Q \lambda_{1q}f_q + e_1\\
x_2 = \hspace{12.5mm} \sum_{q=1}^Q \lambda_{2q}f_q + e_2,
\end{array}
\right., 
\end{eqnarray}
where $\lambda_{1q}\lambda_{2q} \neq 0$ due to the definition of latent confounders, that is, that they contribute to determining the values of more than two variables. 

An estimation method based on overcomplete ICA \citep{Lewicki00NECO} explicitly modeling all the latent confounders $f_q$ was proposed \citep{Hoyer08IJAR}.
However, {\color{black} in current practice,} overcomplete ICA estimation algorithms often get stuck in local optima and are not sufficiently reliable \citep{Entner10AMBN}.
A Bayesian approach for estimating the latent~variable~LiNGAM in Eq.~(\ref{eq:lvlingam}) has been proposed in \citet{Henao11JMLR}. 
These previous approaches that explicitly model latent confounders \citep{Hoyer08IJAR,Henao11JMLR} need to select the number of latent confounders, and which can be quite large. This could lead to further computational difficulty and {\color{black} statistically} unreliable estimates. 

In \citet{Chen13NECO}, a simple approach based on fourth-order cumulants for estimating latent~variable~LiNGAM was proposed. 
Their approach does not need to explicitly model the latent confounders, however it requires the latent confounders $f_q$ to be Gaussian.
The development of nonlinear methods that incorporate latent confounders is ongoing \citep{Zhang10UAI-GP}.

None of these latent confounder methods incorporate the individual-specific effects that we model in the next section to consider latent confounders $f_q$ in the latent variable LiNGAM of Eq.~(\ref{eq:lvlingam}). 

\section{Linear non-Gaussian acyclic structural equation model with individual-specific effects}\label{sec:proposal}
In this section, we propose a new Bayesian method for learning the possible causal direction of two observed variables in the presence of latent confounding variables, assuming that the causal relations are acyclic, i.e., there is not a feedback relation.

\subsection{Model}\label{sec:model}
The LiNGAM \citep{Shimizu06JMLR} for observation $i$ can be described as follows:
\begin{eqnarray}
x_l^{(i)} &=& \mu_l + \sum_{k(m)<k(l)} b_{lm}x_m^{(i)} + e_l^{(i)}. \label{eq:lingam}
\end{eqnarray}
The random variables $e_l^{(i)}$ are non-Gaussian and independent.
The distributions of $e_l^{(i)}$ ($i=1,\cdots,n$) are commonly assumed to be identical\footnote{Relaxing this identically distributed assumption would lead to more general modeling of individual differences, however, this goes beyond the scope of the paper. } for every $l$.
A linear non-Gaussian acyclic structural equation model with individual-specific effects for observation $i$ is formulated as follows:
\begin{eqnarray}
x_l^{(i)} &=& \mu_l + \tilde{\mu}_l^{(i)} + \sum_{k(m)<k(l)} b_{lm}x_m^{(i)} + e_l^{(i)}, \label{eq:model}
\end{eqnarray}
where the difference from LiNGAM is the existence of individual-specific effects $\tilde{\mu}_l^{(i)}$. 
The parameters $\tilde{\mu}_l^{(i)}$ are independent of $e_l^{(i)}$ and are correlated with $x_l^{(i)}$ through the structural equations in our Bayesian approach, introduced below. 
This means that the observations are generated from the identifiable LiNGAM, possibly with different parameter values of the means $\mu_l + \tilde{\mu}_l^{(i)}$. 
We call this a mixed-LiNGAM, named after mixed models \citep{Demidenko04book}, as it has effects {\color{black} $\mu_l$ and $b_{lm}$} that are common to all the observations and individual-specific effects $\tilde{\mu}_l^{(i)}$. 
We note that causal orderings of variables $k(l)$ {\color{black} ($l=1, \cdots, d$)} are identical for all the observations in the sample. 

To use a Bayesian approach for estimating the mixed-LiNGAM, we need to model the distributions of error variables $e_l$ and prior distributions of the parameters including individual-specific effects $\tilde{\mu}_l^{(i)}$, unlike previous LiNGAM methods \citep{Shimizu06JMLR,Hoyer08IJAR}. 
These individual-specific effects, whose number is of the same order as the sample size, are integrated out in the Bayesian method developed in Section~\ref{sec:est}, assuming an informative prior for them {\color{black} similar} to the estimation of conventional mixed models \citep{Demidenko04book}. 
More details on the distributions of error variables and prior distributions of parameters are given in Section~\ref{sec:est}. 
These distributional assumptions were implied to be robust to some extent to their violations, at least in the artificial data experiments of Section~\ref{sec:exp}. 

We now relate the mixed-LiNGAM model above with the latent~variable~LiNGAM \citep{Hoyer08IJAR}. 
The latent~variable~LiNGAM in Eq.~(\ref{eq:lvlingam}) for observation $i$ is written as follows:
\begin{eqnarray}
x_l^{(i)} &=& \mu_l + \sum_{k(m)<k(l)} b_{lm}x_m^{(i)} + \underbrace{\sum_{q=1}^Q \lambda_{lq}f_q^{(i)}}_{\tilde{\mu}_l^{(i)}} + e_l^{(i)}. 
\end{eqnarray}
This is a mixed-LiNGAM taking $\tilde{\mu}_l^{(i)}=\sum_{q=1}^Q\lambda_{lq}f_q^{(i)}$. 
In contrast to the previous approaches for latent~variable~LiNGAM \citep{Hoyer08IJAR,Henao11JMLR}, we do not explicitly model the latent confounders $f_q$ and rather simply include their sums $\tilde{\mu}_l^{(i)}=\sum_{q=1}^Q\lambda_{lq}f_q^{(i)}$ in our model as its parameters since our main interest lies in estimation of the causal relation of observed variables $x_l$ and not in the estimation of their relations with latent confounders $f_q$. 
Our method does not estimate $\lambda_{lq}$ or the number of latent confounders $Q$. 

\subsection{Estimation of possible causal direction}\label{sec:est}
We apply a Bayesian approach to estimate the possible causal direction of two observed variables using the mixed-LiNGAM proposed above.
We compare the following two mixed-LiNGAM models with opposite possible directions of causation.
Model~1 is
\begin{eqnarray}
x_1^{(i)} &=& \mu_1 + \tilde{\mu}_1^{(i)} + e_1^{(i)}\\
x_2^{(i)} &=& \mu_2 + \tilde{\mu}_2^{(i)} + b_{21}x_1^{(i)}
+ e_2^{(i)},
\end{eqnarray}
where $b_{21}$ is non-zero. 
In Model~1, $x_2$ does not cause $x_1$.
The second model, Model~2, is
\begin{eqnarray}
x_1^{(i)} &=& \mu_1 + \tilde{\mu}_1^{(i)} + b_{12}x_2^{(i)}
+ e_1^{(i)}\\
x_2^{(i)} &=& \mu_2 + \tilde{\mu}_2^{(i)} + e_2^{(i)}, 
\end{eqnarray}
where $b_{12}$ is non-zero. 
In Model~2, $x_1$ does not cause $x_2$.
The two models have the same number of parameters, but opposite possible directions of causation. 

Once the possible causal direction is estimated, one can see if the common causal coefficient (connection strength) $b_{21}$ or $b_{12}$ is likely to be zero by examining its posterior distribution.\footnote{\citet{Chickering96AAAI} considered a discrete variable model with {\it known} possible causal direction and proposed a Bayesian approach for computing the posterior distributions of causal effects in the presence of latent confounders.}
We focus here on estimating the possible direction of causation as in many previous works \citep{Dodge01AmericanStatistician,Hoyer09NIPS,Zhang09UAI,Chen13NECO,Hyva13JMLR}, and do not go to the computation of the posterior distribution\footnote{Point estimates of the parameters including the common causal connection strengths $b_{12}$ and $b_{21}$ can be obtained by taking their posterior means based on their posterior distributions, for example.} since estimation of the possible causal direction of two observed variables in the presence of latent confounders has been a very challenging problem in causal inference and is the main topic of this paper. 

{\color{black} We apply standard Bayesian model selection techniques to help assess the} causal direction of $x_1$ and $x_2$.
We use the log-marginal likelihood for comparing the two models. 
{\color{black} The model with the larger log-marginal likelihood is regarded as the closest to the true model} \citep{Kass95JASA}. 

Let ${\mathcal D}$ be the observed dataset $[{\bfx^{(1)}}^T, \cdots, {\bfx^{(n)}}^T]^T$, where $\bfx^{(i)}=[x_1^{(i)}, x_2^{(i)}]^T$.
Denote Models 1 and 2 by $M_1$ and $M_2$.
The log-marginal likelihoods of $M_1$ and $M_2$ are
\begin{eqnarray} 
\log\{ p(M_{\color{black}r} | {\mathcal D}) \} &=& \log\{ p({\mathcal D}|M_{\color{black}r})p(M_{\color{black}r})/p({\mathcal D})\}\\
&=& \log\{ p({\mathcal D}|M_{\color{black}r})\} + \log\{ p(M_{\color{black}r}) \} -\log p({\mathcal D})\\
&= & \log \{ \int p( {\mathcal D}| \btheta_{\color{black}r}, M_{\color{black}r}) p( \btheta_{\color{black}r}| M_{\color{black}r}, \beta_{\color{black}r}) d\btheta_{\color{black}r} \} \nonumber \\
& & + \log p(M_{\color{black}r}) -\log p({\mathcal D})\ ({\color{black}r}=1,2),
\end{eqnarray}
where $\beta_1, \beta_2$ are the hyper-parameter vectors regarding the distributions of the parameters $\btheta_1$ and $\btheta_2$, respectively. 
Since the last term $\log p({\mathcal D})$ is constant with respect to $M_{\color{black}r}$, we can drop it. 
To select suitable values for these hyper-parameters, we take an ordinary empirical Bayesian approach.
First, we compute the log-marginal likelihood for every combination of the two models $M_{\color{black}r}$ and a number of candidate hyper-parameter values of $\beta_{\color{black}r}$. Next, we take the model and hyper-parameter values that give the largest log-marginal likelihood, and finally estimate that the model with the largest log-marginal likelihood is better than the other model.

In basic LiNGAM \citep{Shimizu06JMLR}, we have \citep{Hyva10JMLR,Hoyer09UAI}
{\color{black}
\begin{eqnarray}
p(\bfx) &=& \prod_l p_{e_l}\left(x_l - \mu_l - \sum_{k(m)<k(l)} b_{lm} x_m\right).
\end{eqnarray}
}
Thus, in the same manner, the likelihoods under mixed-LiNGAM $p({\mathcal D}|\btheta_{\color{black}r}, M_{\color{black}r})$ (${\color{black}r}=1,2$) are given by
\begin{eqnarray}
p({\mathcal D}|\btheta_{\color{black}r}, M_{\color{black}r}) &=& \Pi_{i=1}^n\ p(\bfx^{(i)}|\btheta_{\color{black}r}, M_{\color{black}r}) \\
&=& \left\{ \begin{array}{l}
\Pi_{i=1}^n\ p_{e_1^{(i)}}(x_1^{(i)}-\mu_1-\tilde{\mu}_1^{(i)} | \btheta_1, M_1)\\
\times \ p_{e_2^{(i)}}(x_2^{(i)} -\mu_2-\tilde{\mu}_2^{(i)} -b_{21}x_1^{(i)} | \btheta_1, M_1)\ \ {\rm for}\ M_1\\
\Pi_{i=1}^n\ p_{e_1^{(i)}}(x_1^{(i)} -\mu_1-\tilde{\mu}_1^{(i)} -b_{12} x_2^{(i)} | \btheta_2, M_2) \\
\times \ p_{e_2^{(i)}}(x_2^{(i)}-\mu_2-\tilde{\mu}_2^{(i)} | \btheta_2, M_2)\ {\rm for}\ M_2
\end{array}\right.
.
\end{eqnarray}

We model the parameters and their prior distributions as follows.\footnote{This is an example. The modeling method could depend on the domain knowledge.}
The prior probabilities of $M_1$ and $M_2$ are uniform:
\begin{eqnarray}
p(M_1) &=& p( M_2 ).
\end{eqnarray}
The distributions of the error variables $e_1^{(i)}$ and $e_2^{(i)}$ are modeled by Laplace distributions with zero mean and variances of ${\rm var}(e_1^{(i)})=h_1^2$ and ${\rm var}(e_2^{(i)})=h_2^2$
as follows:
\begin{eqnarray}
p_{e_1^{(i)}} &=& Laplace( 0, |h_1|/\sqrt{2})\\
p_{e_2^{(i)}} &=& Laplace( 0, |h_2|/\sqrt{2}).
\end{eqnarray}
Here, we simply use a symmetric super-Gaussian distribution, i.e., the Laplace distribution, to model $p_{e_1^{(i)}}$ and $p_{e_2^{(i)}}$, as suggested in \citet{Hyva13JMLR}.
Such super-Gaussian distributions have been reported to often work well in non-Gaussian estimation methods 
including independent component analysis and LiNGAM \citep{Hyva01book,Hyva13JMLR}.
In some cases, a wider class of non-Gaussian distributions might provide a better model for $p_{e_1^{(i)}}$ and $p_{e_2^{(i)}}$, {\it e.g.}, the generalized Gaussian family \citep{Hyva01book}, a finite mixture of Gaussians, or an exponential family distribution combining the Gaussian and Laplace distributions \citep{Hoyer09UAI}.

The parameter vectors $\btheta_1$ and $\btheta_2$ are written as follows:
\begin{eqnarray}
\btheta_1 &=& [\mu_l, b_{21}, h_l, \tilde{\mu}_l^{(i)}]^T \hspace{5mm} (l=1,2; i=1,\cdots, n)\\
\btheta_2 &=& [\mu_l, b_{12}, h_l, \tilde{\mu}_l^{(i)}]^T\hspace{5mm} (l=1,2; i=1,\cdots, n).
\end{eqnarray}

The prior distributions of common effects are Gaussian as follows:
\begin{eqnarray}
\mu_1 &\sim& N( 0, {\color{black}\tau^{cmmn}_{\mu_1}})\\
\mu_2 &\sim& N( 0, {\color{black}\tau^{cmmn}_{\mu_2}})\\
b_{12} &\sim& N( 0, {\color{black}\tau^{cmmn}_{b_{12}}})\\
b_{21} &\sim& N( 0, {\color{black}\tau^{cmmn}_{b_{21}}})\\
h_1 &\sim& N( 0, {\color{black}\tau^{cmmn}_{h_1}})\\
h_2 &\sim& N( 0, {\color{black}\tau^{cmmn}_{h_2}}), 
\end{eqnarray}
where {\color{black} $\tau^{cmmn}_{\mu_1}$, $\tau^{cmmn}_{\mu_2}$, $\tau^{cmmn}_{b_{12}}$, $\tau^{cmmn}_{b_{21}}$, $\tau^{cmmn}_{h_1}$ and $\tau^{cmmn}_{h_2}$ are constants.} 

{\color{black} Generally speaking, we could use various informative prior distributions for the individual-specific effects and then compare candidate priors using the standard model selection approach based on the marginal likelihoods. {\color{black}Below we provide} two examples. }

If the data is generated from a latent~variable~LiNGAM, a special case of mixed-LiNGAM, as shown in Section~\ref{sec:model}, 
the individual-specific effects are the sums of many non-Gaussian independent latent confounders $f_q$ and are dependent.
The central limit theorem states that {\color{black} the sum of independent variables} becomes increasingly close to the Gaussian \citep{Billingsley86book}.
Therefore, in many cases, it could be practical to approximate the non-Gaussian distribution of a variable that is the sum of many non-Gaussian and independent variables by a bell-shaped curve distribution \citep{Sogawa11NN,Chen13NECO}.
This motivates us to model the prior distribution of individual-specific effects by the multivariate $t$-distribution as follows: 
\begin{eqnarray}
\left[
\begin{array}{c}
\tilde{\mu}_1^{(i)} \\
\tilde{\mu}_2^{(i)}
\end{array}
\right]
&=& {\rm diag}\left(\left[ \sqrt{\tau^{indvdl}_1}, \sqrt{\tau^{indvdl}_2}\right]^T\right)\C^{-1/2} \bfu, \label{eq:38}
\end{eqnarray}
where $\tau^{indvdl}_1$ and $\tau^{indvdl}_2$ are constants, $\bfu \sim t_{\nu}( {\bf 0}, \bSigma)$ and $\bSigma=[\sigma_{ab}]$ is a symmetric scale matrix whose diagonal elements are $1$s.
{\color{black} A random variable vector $\bfu$ that follows the multivariate $t$-distribution $t_{\nu}({\bf 0}, \bSigma)$ can be created by $\frac{\bfy}{\sqrt{v/\nu}}$, where $\bfy$ follows the Gaussian distribution $N({\bf 0},\bSigma)$, $v$ follows the chi-squared distribution with $\nu$ degrees of freedom, and $\bfy$ and $v$ are statistically independent {\color{black}\citep{Kotz04book}}. 
Note that $u_i$ have energy correlations \citep{Hyva01NECO}, i.e., correlations of squares ${\rm cov}(u_i^2,u_j^2)>0$ due to the common variable $v$. }
$\C$ is a diagonal matrix whose diagonal elements give the variance of elements of $\bfu$, i.e., $\C=\frac{\nu}{\nu-2}{\rm diag}(\bSigma)$ for $\nu>2$.
The degree of freedom $\nu$ is here taken to be six.
The kurtosis of the univariate Student's $t$-distribution with six degrees of freedom is three, the same as that of the Laplace distribution.

The hyper-parameter vectors $\beta_1$ and $\beta_2$ are
{\color{black}
\begin{eqnarray}
\beta_l &=& [\tau^{cmmn}_{\mu_1}, \tau^{cmmn}_{\mu_2}, \tau^{cmmn}_{b_{12}}, \tau^{cmmn}_{b_{21}}, \tau^{cmmn}_{h_1}, \tau^{cmmn}_{h_2}, \tau^{indvdl}_1, \tau^{indvdl}_2, \sigma_{21}]^T\hspace{2.5mm} (l=1,2).
\end{eqnarray}
}

{\color{black} We want to take the constants $\tau^{cmmn}_{\mu_1}$, $\tau^{cmmn}_{\mu_2}$, $\tau^{cmmn}_{b_{12}}$, $\tau^{cmmn}_{b_{21}}$, $\tau^{cmmn}_{h_1}$ and $\tau^{cmmn}_{h_2}$ to be sufficiently large so that the priors for the common effects are not very informative. 
It depends on the scales of variables when these constants are sufficiently large.
In the experiments in Sections~\ref{sec:exp}--\ref{sec:real}, we set 
$\tau^{cmmn}_{\mu_1} = \tau^{cmmn}_{b_{12}} = \tau^{cmmn}_{h_1} = 10^2$ $\times$ ${\widehat {\rm var}}$ $(x_1)$ and $\tau^{cmmn}_{\mu_2} = \tau^{cmmn}_{b_{21}} = \tau^{cmmn}_{h_2} = 10^2$ $\times$ ${\widehat {\rm var}}$ $(x_2)$ so that they reflect the scales of the corresponding variables.} 

Moreover, 
we take an empirical Bayesian approach for the individual-specific effects. 
We test $\tau^{indvdl}_l=0, 0.2^2 \times {\widehat {\rm var}}(x_l), ..., 0.8^2\times {\widehat {\rm var}}(x_l), 1.0^2\times {\widehat {\rm var}}(x_l)$ ($l=1,2$). That is, we uniformly vary the hyper-parameter value from that with no {\color{black} individual-specific} effects, i.e., 0, to a larger value, i.e., $1.0^2\times {\widehat {\rm var}}(x_l)$, which implies very large individual differences. 
Further, we test $\sigma_{12}=0, \pm 0.3, \pm 0.5, \pm 0.7, \pm 0.9$, i.e., the value with zero correlation and larger values with stronger correlations. 
This means that we test uncorrelated {\color{black} individual-specific} effects as well as correlated ones. 
We take the ordinary Monte Carlo sampling approach to compute the log-marginal likelihoods with 1000 samples for the parameter vectors $\btheta_{\color{black}r}$ (${\color{black}r}=1,2$). 

The assumptions for our model are summarized in Table~\ref{tab:summary}. 
Generally speaking, if the actual probability density function of individual-specific effects is unimodal and most often provides zero or very small absolute values and with few large values, i.e., many of the individual-specific effects are close to zero and many individuals have similar intercepts, the estimation is likely to work. 
If the individuals have very different intercepts, the estimation will not work very well. 

{\color{black}
An alternative way of modeling the prior distribution of individual-specific effects would be to use the multivariate Gaussian distribution as follows: 
\begin{eqnarray}
\left[
\begin{array}{c}
\tilde{\mu}_1^{(i)} \\
\tilde{\mu}_2^{(i)}
\end{array}
\right]
&=& {\rm diag}\left(\left[ \sqrt{\tau^{indvdl}_1}, \sqrt{\tau^{indvdl}_2}\right]^T\right) \bfz, 
\end{eqnarray}
where $\tau^{indvdl}_1$ and $\tau^{indvdl}_2$ are constants, $\bfz \sim N( {\bf 0}, \bSigma)$ and $\bSigma=[\sigma_{ab}]$ is a symmetric scale matrix whose diagonal elements are $1$s. 
Gaussian individual-specific effects or latent confounders would not lead to losing the identifiability \citep{Chen13NECO} since each observation still is generated by the identifiable non-Gaussian LiNGAM. 
{\color{black} However, if errors are Gaussian, there is no guarantee that our method can find correct possible causal direction. 
We could detect their Gaussianity by comparing our mixed-LiNGAM models with Gaussian error models based on their log-marginal likelihoods. 
If the errors are actually Gaussian or close to be Gaussian, Gaussian error models would provide larger log-marginal likelihoods. This would detect situations where our approach cannot find causal direction. }
}

\begin{table}
\begin{center}
\caption{Summary of the assumptions for our mixed-LiNGAM model}
\scriptsize
\begin{tabular}{l}
Model: $x_l^{(i)} = \mu_l + \tilde{\mu}_l^{(i)} + \sum_{k(m)<k(l)} b_{lm}x_m^{(i)} + e_l^{(i)}$ \hspace{1mm} ($l,m=1,2$; $l\neq m$), \\
where $b_{lm}$ are non-zero.\\
\hspace{7mm} $e_l^{(i)}$ ${\color{black}(l=1,2; i=1,\cdots, n})$ are i.i.d..\\ 
\hspace{7mm} $e_l$ ${\color{black}(l=1,2)}$ are mutually independent.\\
\hspace{7mm} $e_l$ ${\color{black}(l=1,2)}$ follow Laplace distributions with zero mean and standard deviations $|h_l|$.\\
\\
\hspace{7mm} Prior distributions: \\
\hspace{7mm} $\mu_l$, $b_{lm}$ and $h_l$ ${\color{black}(l=1,2; m=1,2; l \neq m)}$ follow Gaussian distributions with zero mean and variance {\color{black} $\tau^{cmmn}_{\mu_l}$, }\\
\hspace{7mm} {\color{black} $\tau^{cmmn}_{b_{lm}}$ and $\tau^{cmmn}_{h_l}$.}\\
\hspace{7mm} $\tilde{\mu}_l^{(i)}$ ${\color{black}(l=1,2; i=1, \cdots, n)}$ are the sum of latent confounders $f_q^{(i)}$: $\sum_{q=1}^Q\lambda_{lq}f_q^{(i)}$ and are independent of $e_l^{(i)}$.\\
\hspace{7mm} $\tilde{\mu}_l^{(i)}$ ${\color{black}(l=1,2; i=1, \cdots, n)}$ are i.i.d.. \\
\hspace{7mm} ${\color{black}\mu_l}$ ${\color{black}(l=1,2)}$ follow multivariate $t$-distributions with $\nu$ degrees of freedom, zero mean, variances $\tau_l^{indvdl}$ \\\hspace{7mm} and correlation $\sigma_{12}$ (here, $\nu=6$). \\
\\
\hspace{7mm} Hyper-parameters:\\
\hspace{7mm} {\color{black} $\tau^{cmmn}_{\mu_l}$, $\tau^{cmmn}_{b_{lm}}$ and $\tau^{cmmn}_{h_l}$ ${\color{black}(l=1,2; m=1,2; l \neq m)}$} are set to be large values so that the priors are not\\
\hspace{7mm} very informative.\\
\hspace{7mm} $\tau_l^{indvdl}$ ${\color{black}(l=1,2)}$ are uniformly varied from zero to large values.\\
\hspace{7mm} $\sigma_{12}$ are uniformly varied in the interval between -0.9 and 0.9.\\
\end{tabular}
\label{tab:summary}
\end{center}
\end{table}

\section{Experiments on artificial data}\label{sec:exp}
We compared our method with {\color{black} seven} methods for estimating the possible causal direction between two variables: 
{\color{black}
i) LvLiNGAM\footnote{\label{footnote:LvLiNGAM}\url{http://www.cs.helsinki.fi/u/phoyer/code/lvlingam.tar.gz}} \citep{Hoyer08IJAR}; 
ii) SLIM\footnote{\url{http://cogsys.imm.dtu.dk/slim/}}\citep{Henao11JMLR}
iii) LiNGAM-GC-UK~\citep{Chen13NECO}; 
}
iv) ICA-LiNGAM\footnote{\url{http://www.cs.helsinki.fi/group/neuroinf/lingam/lingam.tar.gz}}~\citep{Shimizu06JMLR}; 
v) DirectLiNGAM\footnote{\url{http://www.ar.sanken.osaka-u.ac.jp/~sshimizu/code/Dlingamcode.html}} \citep{Shimizu11JMLR}; 
vi) Pairwise~LiNGAM\footnote{\url{http://www.cs.helsinki.fi/u/ahyvarin/code/pwcausal/}}~\citep{Hyva13JMLR}; 
vii) Post-nonlinear causal model (PNL) \footnote{\url{http://webdav.tuebingen.mpg.de/causality/CauseOrEffect_NICA.rar}} \citep{Zhang09UAI}. 
Their assumptions are summarized in Table~\ref{tab:summary2}. 
The first {\color{black}seven} methods assume linearity, and the {\color{black}eighth} allows a very wide variety of nonlinear relations. The last four methods assume that there are no latent confounders. 
{\color{black} 
We tested the prior $t$- and Gaussian distributions for individual-specific effects in our approach. LvLiNGAM and SLIM require to specify the number of latent confounders. We tested 1 and 4 latent confounder(s) for LvLiNGAM since its current implementation cannot handle more than four latent confounders, whereas we tested 1, 4 and 10 latent confounders(s) for SLIM. 
LiNGAM-GC-UK \citep{Chen13NECO} assumes that errors are simultaneously super-Gaussian or sub-Gaussian and that latent confounders are Gaussian.}

\begin{table}[]
\begin{center}
\caption{Summary of the assumptions of {\color{black}eight} methods}
\scriptsize
\begin{tabular}{l|llllll}
& Functional & Latent & {\color{black} Number of} & Iterative search & Distributional & \\
& form? & confounders & {\color{black} latent confounders} & in the parameter & assumptions &\\
& & allowed? & {\color{black} necessary} & space required? & necessary?\\
& & & {\color{black} to be specified?} & & \\
\hline
Our approach & Linear & Yes & {\color{black} No} & No & Yes \\
{\color{black} LvLiNGAM} & {\color{black} Linear} & {\color{black} Yes} & {\color{black} Yes} & {\color{black} Yes} & {\color{black} No\footnotemark}\\
{\color{black} SLIM} & {\color{black} Linear} & {\color{black} Yes} & {\color{black} Yes} & {\color{black} No} & {\color{black} Yes}\\
{\color{black} LiNGAM-GC-UK} & {\color{black} Linear} & {\color{black} Yes} & {\color{black} No} & {\color{black} No} & {\color{black} Yes}\\
ICA-LiNGAM & Linear & No & {\color{black} N/A} & Yes & No \\
DirectLiNGAM & Linear & No & {\color{black} N/A} & No & No \\ 
Pairwise~LiNGAM & Linear & No & {\color{black} N/A} & No & No \\
PNL & Nonlinear & No & {\color{black} N/A} & Yes & No
\end{tabular}
\label{tab:summary2}
\end{center}
\end{table}
\footnotetext{{\color{black} Their current implementation of LvLiNGAM in Footnote~\ref{footnote:LvLiNGAM} assumes a non-Gaussian distribution, which is a mixture of two Gaussian distributions. }}

\begin{figure}[t]
\begin{center}
\includegraphics[width=0.2\textwidth]{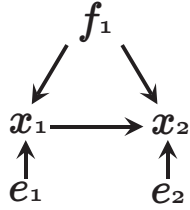}
\caption{The associated graph of the model used to generate artificial data when the number of latent confounders $Q=1$.}
\label{fig:example2}
\end{center}
\end{figure}

We generated data using the following latent~variable~LiNGAM with $Q$ latent confounding variables, which is a mixed-LiNGAM:
\begin{eqnarray}
x_1^{(i)} &=& \mu_1 + \sum_{q=1}^Q\lambda_{1q}f_q^{(i)} + e_1^{(i)}\\
x_2^{(i)} &=& \mu_2 + b_{21} x_1^{(i)} + \sum_{q=1}^Q \lambda_{2q}f_q^{(i)} + e_2^{(i)}, \label{eq:setup3}
\end{eqnarray}
where $\mu_1$ and $\mu_2$ were randomly generated from $N(0,1)$,
and $b_{21}, \lambda_{1q}, \lambda_{2q}$ were randomly generated from the interval $(-1.5,-0.5) \cup (0.5,1.5)$. 
We tested various numbers of latent confounders $Q=0, 1, 6, 12$. 
The zero values indicate that there are no latent confounders. 
An example graph used to generate artificial data is given in Fig.~\ref{fig:example2}. 

The distributions of the error variables $e_1$, $e_2$, and latent confounders $f_q$ were identical for all observations.
The distributions of the error variables $e_1$, $e_2$, and latent confounders $f_q$ were randomly selected from the 18 non-Gaussian distributions used in \citet{Bach02JMLR} to see if the Laplace distribution assumption on error variables and $t$- {\color{black} or Gaussian} distribution assumption on individual-specific effects in our method were robust to different non-Gaussian distributions. These include symmetric/non-symmetric distributions, super-Gaussian/sub-Gaussian distributions, 
and strongly/weakly non-Gaussian distributions. 
The variances of $e_1$ and $e_2$ were randomly selected from the interval $(0.5^2,1.5^2)$.
The variances of $f_q$ were {\color{black}1s}.

We permuted the variables according to a random ordering and inputted them to the {\color{black} eight} estimation methods. We conducted 100 trials,
with sample sizes of 50, 100, and 200. 
For the data with the number of latent confounders $Q=0$, all the methods should find the correct causal direction for large enough sample sizes, as there were no latent confounders, which here means no individual-specific effects.
The {\color{black} last four} comparative methods should find the data with the number of latent confounders $Q=1, 6, 12$ very difficult to analyze, because, unlike the other approaches, they assume no {\color{black} latent confounders}.

To evaluate the performance of the algorithms, we counted the number of successful discoveries of possible causal direction {\color{black} and estimated their standard errors. }

{\color{black} Looking at Table~\ref{sim:success} as a whole there are several general observations that we can make. First though none of the procedures is infallible, several of them do quite well in that they choose the correct causal direction about 90\% of the time. Second, overall our approach is the most successful across the conditions of the simulation. Specifically, in all but the cases of no confounding variables, one or both of our approaches have the highest percentages of success. In the situation of no confounding variables, ICA-LiNGAM, DirectLiNGAM, and Pairwise~LINGAM have higher success percentages than our procedures. 
These generalizations need qualifications in that there are sampling errors that affect the estimates. Formal tests of significance across all conditions would be complicated. It would require taking account of multiple testing and the dependences of the simulated samples under the same sample size and number of confounders. However, the standard errors of the estimated percentages serve to caution the reader not to judge the percentages alone without recognizing sampling variability. For instance, when there are no confounders and a sample size of 50, the ICA-LiNGAM procedure appears best with 93\% success, but the success percentages of our two approaches fall within two standard errors of the 93\% estimate. Alternatively, in the rows with 6 confounders and sample size 50 our approach with 88\% success and a standard error of 3.25 appears sufficiently far from the success percentages of the other methods besides ours to make sampling fluctuations an unlikely explanation. 
In sum, taking all the evidence together, our approaches performed quite well and deserve further investigation under additional simulation conditions. 
}

Table~\ref{sim:time} shows the average computational times.
The computational complexity of the current implementation of {\color{black} our methods} is clearly larger than that of the other linear methods ICA-LiNGAM, DirectLiNGAM, Pairwise~LiNGAM, {\color{black} LvLiNGAM with 1 latent confounder, SLIM and LiNGAM-GC-UK} and comparable to {\color{black} LvLiNGAM with 4 latent confounders} and the nonlinear method PNL.

The MATLAB code for performing these experiments is available on our website.\footnote{\url{http://www.ar.sanken.osaka-u.ac.jp/~sshimizu/code/mixedlingamcode.html}}

\begin{table}[ht]
\begin{center}
\caption{Number of successful discoveries (100 trials)}
\scriptsize
\begin{tabular}{l|rrr}
& \multicolumn{3}{c}{Sample size} \\
& 50 & 100 & 200\\ \hline \hline
Number of latent confounders $Q=0$: & & \\ 
\hspace{1.5mm} Our approach ($t$-distributed individual-specific effects) & 88 {\color{black} (3.25)} & 91 {\color{black} (2.86)} & 86 {\color{black} (3.47)} \\
\hspace{1.5mm} {\color{black} Our approach (Gaussian individual-specific effects)} & 91 {\color{black} (2.86)} & 87 {\color{black} (3.36)} & 91 {\color{black} (2.86)} \\
\hspace{1.5mm} {\color{black} LvLiNGAM (1 latent confounder)} & 73 {\color{black} (4.44)} & 83 {\color{black} (3.76)} & 83 {\color{black} (3.76)} \\
\hspace{1.5mm} {\color{black} LvLiNGAM (4 latent confounders)} & 52 {\color{black} (5.00)} & 68 {\color{black} (4.66)} & 66 {\color{black} (4.74)} \\
\hspace{1.5mm} {\color{black} SLIM (1 latent confounder)} & 29 {\color{black} (4.54)} & 30 {\color{black} (4.58)} & 25 {\color{black} (4.33)} \\
\hspace{1.5mm} {\color{black} SLIM (4 latent confounders)} & 34 {\color{black} (4.74)} & 31 {\color{black} (4.62)} & 36 {\color{black} (4.80)} \\
\hspace{1.5mm} {\color{black} SLIM (10 latent confounders)} & 30 {\color{black} (4.58)} & 29 {\color{black} (4.54)} & 30 {\color{black} (4.58)} \\
\hspace{1.5mm} {\color{black} LiNGAM-GC-UK} & 33 {\color{black} (4.70)} & 28 {\color{black} (4.49)} & 35 {\color{black} (4.77)} \\
\hspace{1.5mm} ICA-LiNGAM & \underline{93} {\color{black} (2.55)} & 93 {\color{black} (2.55)} & 96 {\color{black} (1.96)} \\
\hspace{1.5mm} DirectLiNGAM & 87 {\color{black} (3.36)} & \underline{95} {\color{black} (2.18)} & \underline{97} {\color{black} (1.71)} \\
\hspace{1.5mm} Pairwise~LiNGAM & 89 {\color{black} (3.13)} & \underline{95} {\color{black} (2.18)} & 95 {\color{black} (2.18)} \\ 
\hspace{1.5mm} Post-nonlinear causal model & 74 {\color{black} (4.39)} & 71 {\color{black} (4.54)} & 75 {\color{black} (4.33)} \\
\hline
Number of latent confounders $Q=1$: & & \\ 
\hspace{1.5mm} Our approach {\color{black}($t$-distributed individual-specific effects)}& \underline{83} {\color{black} (3.76)} & 80 {\color{black} (4.00)} & \underline{80} {\color{black} (4.00)} \\
\hspace{1.5mm} {\color{black} Our approach (Gaussian individual-specific effects)} & 79 {\color{black} (4.07)} & \underline{87} {\color{black} (3.36)} & 69 {\color{black} (4.62)} \\
\hspace{1.5mm} {\color{black} LvLiNGAM (1 latent confounder)} & 66 {\color{black} (4.74)} & 71 {\color{black} (4.54)} & 73 {\color{black} (4.44)} \\
\hspace{1.5mm} {\color{black} LvLiNGAM (4 latent confounders)} & 63 {\color{black} (4.83)} & 58 {\color{black} (4.94)} & 67 {\color{black} (4.70)} \\
\hspace{1.5mm} {\color{black} SLIM (1 latent confounder)} & 40 {\color{black} (4.90)} & 47 {\color{black} (4.99)} & 25 {\color{black} (4.33)} \\
\hspace{1.5mm} {\color{black} SLIM (4 latent confounders)} & 40 {\color{black} (4.90)} & 34 {\color{black} (4.74)} & 44 {\color{black} (4.96)} \\
\hspace{1.5mm} {\color{black} SLIM (10 latent confounders)} & 47 {\color{black} (4.99)} & 39 {\color{black} (4.88)} & 41 {\color{black} (4.92)}\\
\hspace{1.5mm} {\color{black} LiNGAM-GC-UK} & 24 {\color{black} (4.27)} & 32 {\color{black} (4.66)} & 32 {\color{black} (4.66)}\\
\hspace{1.5mm} ICA-LiNGAM & 74 {\color{black} (4.39)} & 71 {\color{black} (4.54)} & 67 {\color{black} (4.70)}\\
\hspace{1.5mm} DirectLiNGAM & 48 {\color{black} (5.00)} & 52 {\color{black} (5.00)} & 54 {\color{black} (4.98)}\\
\hspace{1.5mm} Pairwise~LiNGAM & 54 {\color{black} (4.98)} & 58 {\color{black} (4.94)} & 61 {\color{black} (4.88)}\\ 
\hspace{1.5mm} Post-nonlinear causal model & 55 {\color{black} (4.97)} & 58 {\color{black} (4.94)} & 57 {\color{black} (4.95)}\\
\hline
Number of latent confounders $Q=6$: & & & \\
\hspace{1.5mm} Our approach ($t$-distributed individual-specific effects)& \underline{88} {\color{black} (3.25)} & 81 {\color{black} (3.92)} & \underline{87} {\color{black} (3.36)}\\
\hspace{1.5mm} {\color{black} Our approach (Gaussian individual-specific effects)} & 84 {\color{black} (3.67)} & \underline{85} {\color{black} (3.57)} & \underline{87} {\color{black} (3.36)} \\
\hspace{1.5mm} {\color{black} LvLiNGAM (1 latent confounder)} & 58 {\color{black} (4.94)} & 70 {\color{black} (4.58)} & 70 {\color{black} (4.58)}\\
\hspace{1.5mm} {\color{black} LvLiNGAM (4 latent confounders)} & 64 {\color{black} (4.80)} & 61 {\color{black} (4.88)} & 63 {\color{black} (4.83)}\\
\hspace{1.5mm} {\color{black} SLIM (1 latent confounder)} & 50 {\color{black} (5.00)} & 63 {\color{black} (4.83)} & 47 {\color{black} (4.99)}\\
\hspace{1.5mm} {\color{black} SLIM (4 latent confounders)} & 45 {\color{black} (4.97)}& 47 {\color{black} (4.99)} & 43 {\color{black} (4.95)}\\
\hspace{1.5mm} {\color{black} SLIM (10 latent confounders)} & 58 {\color{black} (4.94)} & 48 {\color{black} (5.00)} & 58 {\color{black} (4.94)}\\
\hspace{1.5mm} {\color{black} LiNGAM-GC-UK} & 29 {\color{black} (4.54)} & 28 {\color{black} (4.49)} & 21 {\color{black} (4.07)}\\
\hspace{1.5mm} ICA-LiNGAM & 74 {\color{black} (4.39)} & 72 {\color{black} (4.49)} & 47 {\color{black} (4.99)}\\
\hspace{1.5mm} DirectLiNGAM & 37 {\color{black} (4.83)} & 48 {\color{black} (5.00)} & 39 {\color{black} (4.88)}\\
\hspace{1.5mm} Pairwise~LiNGAM & 48 {\color{black} (5.00)} & 51 {\color{black} (5.00)} & 37 {\color{black} (4.83)}\\
\hspace{1.5mm} Post-nonlinear causal model & 55 {\color{black} (4.97)} & 42 {\color{black} (4.94)} & 46 {\color{black} (4.98)}\\
\hline
Number of latent confounders $Q=12$: & & \\
\hspace{1.5mm} Our approach ($t$-distributed individual-specific effects)& 88 {\color{black} (3.25)} & 86 {\color{black} (3.47)} & 89 {\color{black} (3.13)}\\
\hspace{1.5mm} {\color{black} Our approach (Gaussian individual-specific effects)} & \underline{91} {\color{black} (2.86)} & \underline{89} {\color{black} (3.13)} & \underline{91} {\color{black} (2.86)}\\
\hspace{1.5mm} {\color{black} LvLiNGAM (1 latent confounder)} & 52 {\color{black} (5.00)} & 55 {\color{black} (4.97)} & 65 {\color{black} (4.77)}\\
\hspace{1.5mm} {\color{black} LvLiNGAM (4 latent confounders)} & 65 {\color{black} (4.77)} & 58 {\color{black} (4.94)} & 64 {\color{black} (4.80)}\\
\hspace{1.5mm} {\color{black} SLIM (1 latent confounder)} & 51 {\color{black} (5.00)} & 55 {\color{black} (4.97)} & 60 {\color{black} (4.90)}\\
\hspace{1.5mm} {\color{black} SLIM (4 latent confounders)} & 45 {\color{black} (4.97)} & 51 {\color{black} (5.00)} & 63 {\color{black} (4.83)}\\
\hspace{1.5mm} {\color{black} SLIM (10 latent confounders)} & 61 {\color{black} (4.88)} & 54 {\color{black} (4.98)} & 54 {\color{black} (4.98)} \\
\hspace{1.5mm} {\color{black} LiNGAM-GC-UK} & 21 {\color{black} (4.07)} & 25 {\color{black} (4.33)} & 29 {\color{black} (4.54)} \\
\hspace{1.5mm} ICA-LiNGAM & 68 {\color{black} (4.66)} & 72 {\color{black} (4.49)} & 72 {\color{black} (4.49)}\\
\hspace{1.5mm} DirectLiNGAM & 37 {\color{black} (4.83)} & 39 {\color{black} (4.88)} & 38 {\color{black} (4.85)} \\
\hspace{1.5mm} Pairwise~LiNGAM & 56 {\color{black} (4.96)} & 42 {\color{black} (4.94)} & 43 {\color{black} (4.95)} \\ 
\hspace{1.5mm} Post-nonlinear causal model & 51 {\color{black} (5.00)} & 43 {\color{black} (4.95)} & 46 {\color{black} (4.98)}\\
\multicolumn{4}{l}{}\\
\multicolumn{4}{l}{Largest numbers of successful discoveries were underlined. }\\
\multicolumn{4}{l}{{\color{black} Standard errors are shown in parentheses, which are computed assuming that the number}}\\
\multicolumn{4}{l}{{\color{black}of successes follow a binomial distribution.}}\\
\end{tabular}
\label{sim:success}
\end{center}
\end{table}%

\begin{table}[ht]
\begin{center}
\caption{Average CPU time (s)}
\scriptsize
\begin{tabular}{l|rrr}
& \multicolumn{3}{c}{Sample size} \\
& 50 & 100 & 200 \\ \hline \hline
Number of latent confounders $Q=0$ & & & \\
\hspace{1.5mm} Our approach ($t$-distributed individual-specific effects) & {\color{black} 27.20} & {\color{black} 56.93} & {\color{black} 141.84} \\
\hspace{1.5mm} {\color{black} Our approach (Gaussian individual-specific effects)} & 35.48 & 69.59 & 117.10 \\
\hspace{1.5mm} {\color{black} LvLiNGAM (1 latent confounder)} & 2.41 & 2.55 & 9.91 \\
\hspace{1.5mm} {\color{black} LvLiNGAM (4 latent confounders)} & 22.25 & 30.12 & 87.96 \\
\hspace{1.5mm} {\color{black} SLIM (1 latent confounder)} & 5.89 & 6.25 & 6.81\\
\hspace{1.5mm} {\color{black} SLIM (4 latent confounders)} & 7.60 & 8.14 & 9.13 \\
\hspace{1.5mm} {\color{black} SLIM (10 latent confounders)} & 10.88 & 12.02 & 13.96 \\
\hspace{1.5mm} {\color{black} LiNGAM-GC-UK} & 0.00 & 0.00 & 0.00 \\
\hspace{1.5mm} ICA-LiNGAM & 0.04 & 0.03 & 0.02 \\
\hspace{1.5mm} DirectLiNGAM & 0.00 & 0.01 & 0.01 \\
\hspace{1.5mm} Pairwise~LiNGAM & 0.00 & 0.00 & 0.00 \\ 
\hspace{1.5mm} Post-nonlinear causal model & 19.59 & 27.68 & 57.37 \\
\hline
Number of latent confounders $Q=1$: & & \\
\hspace{1.5mm} Our approach ($t$-distributed individual-specific effects) & {\color{black} 35.87} & {\color{black} 65.55} & {\color{black} 131.25}\\
\hspace{1.5mm} {\color{black} Our approach (Gaussian individual-specific effects)} & 37.12 & 75.11 & 114.37 \\
\hspace{1.5mm} {\color{black} LvLiNGAM (1 latent confounder)} & 2.40 & 2.53 & 13.93\\
\hspace{1.5mm} {\color{black} LvLiNGAM (4 latent confounders)} & 21.50 & 29.50 & 92.19\\
\hspace{1.5mm} {\color{black} SLIM (1 latent confounder)} & 5.88 & 6.01 & 6.69\\
\hspace{1.5mm} {\color{black} SLIM (4 latent confounders)} & 7.59 & 8.19 & 8.96 \\
\hspace{1.5mm} {\color{black} SLIM (10 latent confounders)} & 10.96 & 11.79 & 13.68 \\
\hspace{1.5mm} {\color{black} LiNGAM-GC-UK} & 0.00 & 0.00 & 0.00 \\
\hspace{1.5mm} ICA-LiNGAM & 0.05 & 0.03 & 0.03 \\
\hspace{1.5mm} DirectLiNGAM & 0.01 & 0.01 & 0.01 \\
\hspace{1.5mm} Pairwise~LiNGAM & 0.00 & 0.00 & 0.00 \\ 
\hspace{1.5mm} Post-nonlinear causal model & 18.17 & 28.83 & 51.63 \\
\hline
Number of latent confounders $Q=6$: & & & \\
\hspace{1.5mm} Our approach ($t$-distributed individual-specific effects) & {\color{black} 42.66} & {\color{black} 76.29} & {\color{black} 132.43}\\
\hspace{1.5mm} {\color{black} Our approach (Gaussian individual-specific effects)} & 33.13 & 69.07 & 104.83 \\
\hspace{1.5mm} {\color{black} LvLiNGAM (1 latent confounder)} & 2.40 & 2.56 & 9.38 \\
\hspace{1.5mm} {\color{black} LvLiNGAM (4 latent confounders)} & 22.17 & 30.12 & 83.01 \\
\hspace{1.5mm} {\color{black} SLIM (1 latent confounder)} & 5.89 & 6.22 & 6.77 \\
\hspace{1.5mm} {\color{black} SLIM (4 latent confounders)} & 7.58 & 8.18 & 9.11 \\
\hspace{1.5mm} {\color{black} SLIM (10 latent confounders)} & 11.03 & 12.02 & 13.91 \\
\hspace{1.5mm} {\color{black} LiNGAM-GC-UK} & 0.00 & 0.00 & 0.00 \\
\hspace{1.5mm} ICA-LiNGAM & 0.06 & 0.05 & 0.05 \\
\hspace{1.5mm} DirectLiNGAM & 0.01 & 0.01 & 0.01 \\
\hspace{1.5mm} Pairwise~LiNGAM & 0.00 & 0.00 & 0.00 \\ 
\hspace{1.5mm} Post-nonlinear causal model & 18.71 & 29.62 & 52.21 \\
\hline
Number of latent confounders $Q=12$: & & \\
\hspace{1.5mm} Our approach ($t$-distributed individual-specific effects) & {\color{black} 29.16} & {\color{black} 59.30} & {\color{black} 134.89} \\
\hspace{1.5mm} {\color{black} Our approach (Gaussian individual-specific effects)} & 32.18 & 68.14 & 104.76 \\
\hspace{1.5mm} {\color{black} LvLiNGAM (1 latent confounder)} & 2.35 & 2.50 & 13.58 \\
\hspace{1.5mm} {\color{black} LvLiNGAM (4 latent confounders)} & 21.51 & 30.10 & 94.08 \\
\hspace{1.5mm} {\color{black} SLIM (1 latent confounder)} & 5.90 & 6.03 & 6.62 \\
\hspace{1.5mm} {\color{black} SLIM (4 latent confounders)} & 7.58 & 7.99 & 8.97 \\
\hspace{1.5mm} {\color{black} SLIM (10 latent confounders)} & 10.92 & 11.68 & 13.74 \\
\hspace{1.5mm} {\color{black}LiNGAM-GC-UK} & 0.00 & 0.00 & 0.00 \\
\hspace{1.5mm} ICA-LiNGAM & 0.07 & 0.08 & 0.07 \\
\hspace{1.5mm} DirectLiNGAM & 0.01 & 0.02 & 0.02 \\
\hspace{1.5mm} Pairwise~LiNGAM & 0.00 & 0.00 & 0.00 \\ 
\hspace{1.5mm} Post-nonlinear causal model & 18.21 & 29.21 & 51.89 
\end{tabular}
\label{sim:time}
\end{center}
\end{table}%

\section{{\color{black} An experiment} on real-world data}\label{sec:real}
We analyzed the General Social Survey data set, taken from a sociological data repository (\url{http://www.norc.org/GSS+Website/}).
The data consisted of six observed variables: $x_1$: {\color{black} prestige of father's occupation}, $x_2$: son's income, $x_3$: father's education, $x_4$: {\color{black} prestige of son's occupation}, $x_5$: son's education, and $x_6$: number of siblings.\footnote{Although $x_6$ is discrete, it can be considered as continuous because it is an ordinal scale with many points.}
The sample selection was conducted based on the following criteria: i) non-farm background; ii) ages 35--44; iii) white; iv) male; v) in the labor force at the time of the survey; vi) not missing data for any of the covariates; and vii) data taken from 1972--2006. The sample size was 1380.

\begin{figure}[t]
\begin{center}
\includegraphics[width=0.6\textwidth]{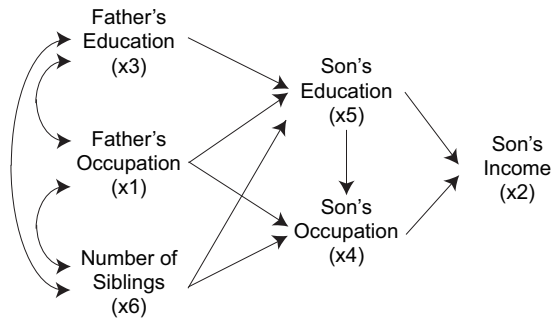}
\caption{Status attainment model based on domain knowledge. Usually, the relations of $x_1$, $x_3$, and $x_6$, represented by bi-directed arcs, are not modeled.}
\label{fig:domain}
\end{center}
\end{figure}

The possible directions were determined based on the domain knowledge in \citet{Duncan72book}, shown in Fig.~\ref{fig:domain}. 
The causal relations of $x_1$, $x_3$, and $x_6$ usually are not modeled in the literature since there are many other determinants of these three exogenous observed variables that are not part of the model. 
{\color{black} However, the possible causal directions among the three variables would be $x_1 \leftarrow x_3$, $x_6 \leftarrow x_1$, and $x_6 \leftarrow x_3$ based on their temporal orders.}

Table~\ref{tab:Bollen} shows {\color{black} the numbers of successes and precisions.}
Our mixed-LiNGAM approach {\color{black} with the $t$-distributed individual-specific effects} gave the largest number of successful discoveries {\color{black} 12} and achieved the highest precision {\color{black}, i.e., num.~successes / num.~pairs = 12/15 = 0.80}. 
The second best method was {\color{black} our mixed-LiNGAM approach with the Gaussian individual-specific effects}, which found {\color{black} one less correct possible directions than the $t$-distribution version}. 
{\color{black} The third best method was LvLiNGAM with 1 latent confounder, which found two less correct possible directions than the $t$-distribution version}. 
This would be mainly because our two methods allow individual-specific effects and the other methods do not. 

{\color{black}Table~\ref{tab:esthyperparameters} shows the estimated hyper-parameter values of our mixed-LiNGAM approach with the $t$-distributed individual-specific effects that performed best in the sociology data experiment.}
Either the estimated hyper-parameter $\hat{\tau}^{indvdl}_1$ or $\hat{\tau}^{indvdl}_2$ that represents the magnitudes of individual differences was non-zero in all pairs except ($x_4, x_5$). 
The non-ignorable influence of latent confounders was implied between the pairs ($x_2, x_4$), ($x_2, x_6$) and ($x_3, x_6$) since both $\hat{\tau}^{indvdl}_1$ or $\hat{\tau}^{indvdl}_2$ were non-zero for the pairs.
In addition, for the pair ($x_2, x_6$), there might exist some nonlinear influence of latent confounders, since $\hat{\sigma}_{12}$ is zero, i.e., the individual-specific effects were linearly uncorrelated but dependent.\footnote{Two variables that follow the multivariate $t$-distribution are dependent, even when they are uncorrelated, {\color{black}as stated in Section~\ref{sec:est}}.}
{\color{black} If $\hat{\sigma}_{12}$ were larger, it would have implied a larger linear influence of the latent confounders on the pair ($x_2, x_6$).}
The estimates of the hyper-parameter $\tau^{indvdl}_1$ were very large for the pairs ($x_2, x_6$) and ($x_4, x_1$), which implied very large individual differences regarding $x_2$ and $x_4$ respectively. This might imply that the estimated directions could be less reliable, although they were correct in this example. 

Another point to note is that {\color{black} both our methods with $t$-distributed and Gaussian individual-specific effects} failed to find the possible direction $x_5 \leftarrow x_1$, although the causal relation is expected to occur from the domain knowledge \citep{Duncan72book}. 
This failure would be attributed to the model misspecification since the sample size was very large.
Since the estimate of the hyper-parameter $\tau^{indvdl}_1$ regarding $x_5$ was zero, the influence of latent confounders might be small for this pair, although the estimate of $\tau^{indvdl}_2$ was not small and the individual difference regarding $x_5$ seemed substantial. 
Modeling both latent confounders and nonlinear relations and/or allowing a wider class of non-Gaussian distributions might lead to better performance. 
This is an important line of future research.

\begin{table}[ht]
\begin{center}
\caption{Comparison of {\color{black}eight} methods}
\scriptsize
\begin{tabular}{l|ccccccc}
Possible directions & \multicolumn{2}{c}{Our approach} & \multicolumn{2}{c}{{\color{black} LvLiNGAM}} & \multicolumn{3}{c}{{\color{black} SLIM}} \\
& $t$-dist. & {\color{black}Gaussian} & \multicolumn{2}{c}{{\color{black} Num. lat. conf.}} & \multicolumn{3}{c}{{\color{black} Num. lat. conf.}}\\
& & & 1 & 4 & 1 & 4 &10\\
{\color{black} $x_1 (FO) \leftarrow x_3 (FE)$} & \checkmark & \checkmark & & \checkmark & & & \checkmark \\
$x_2 (SI) \leftarrow x_1 (FO) $ & \checkmark & \checkmark & & & & \checkmark & \checkmark \\
$x_2 (SI) \leftarrow x_3 (FE) $ & \checkmark & \checkmark & & & \checkmark & \checkmark & \checkmark \\
$x_2 (SI) \leftarrow x_4 (SO) $ & \checkmark & \checkmark & & & \checkmark & \checkmark & \\
$x_2 (SI) \leftarrow x_5 (SE)$ & \checkmark & \checkmark & \checkmark & \checkmark & \checkmark & & \checkmark \\
$x_2 (SI) \leftarrow x_6 (NS)$ & \checkmark & \checkmark & \checkmark & \checkmark & & & \\
$x_4 (SO) \leftarrow x_1 (FO) $ & \checkmark & \checkmark & \checkmark & \checkmark & \checkmark & \checkmark & \checkmark \\
$x_4 (SO) \leftarrow x_3 (FE) $ & \checkmark & \checkmark & \checkmark & & \checkmark & \checkmark & \checkmark\\
$x_4 (SO) \leftarrow x_5 (SE)$ & \checkmark & \checkmark & \checkmark & \checkmark & & & \\
$x_4 (SO) \leftarrow x_6 (NS) $ & \checkmark & \checkmark & \checkmark & \checkmark & \checkmark & & \\
$x_5 (SE) \leftarrow x_1 (FO)$ & & & & & \checkmark & & \\
$x_5 (SE) \leftarrow x_3 (FE) $ & \checkmark & & \checkmark & \checkmark & \checkmark & \checkmark & \\
$x_5 (SE) \leftarrow x_6 (NS) $ & \checkmark & \checkmark & \checkmark & \checkmark & \checkmark & & \\
{\color{black} $x_6 (NS) \leftarrow x_1 (FO)$} & & & \checkmark & & & & \checkmark \\
{\color{black} $x_6 (NS) \leftarrow x_3 (FE)$} & & & \checkmark & \checkmark & & \checkmark & \checkmark\\
\hline
Num. of successes & 12 & 11 & 10 & 9 & 9 & 7 & 8 \\
Precisions & 0.80 & 0.73 & 0.67 & 0.60 & 0.60 & 0.47 & 0.53 \\
\hline
Possible directions & {\color{black} LiNGAM-GC-UK} & ICA & Direct & Pairwise & PNL & \multicolumn{2}{c}{}\\
{\color{black} $x_1 (FO) \leftarrow x_3 (FE)$} & & \checkmark & \checkmark & & \multicolumn{2}{c}{}\\
$x_2 (SI) \leftarrow x_1 (FO) $ & & \checkmark & \checkmark & & \checkmark & \multicolumn{2}{c}{}\\
$x_2 (SI) \leftarrow x_3 (FE) $ & & \checkmark & & & \checkmark & \multicolumn{2}{c}{}\\
$x_2 (SI) \leftarrow x_4 (SO) $ & & \checkmark & \checkmark & & \checkmark & \multicolumn{2}{c}{}\\
$x_2 (SI) \leftarrow x_5 (SE)$ & & \checkmark & & & \checkmark & \multicolumn{2}{c}{}\\
$x_2 (SI) \leftarrow x_6 (NS)$ & & \checkmark & & & \checkmark & \multicolumn{2}{c}{}\\
$x_4 (SO) \leftarrow x_1 (FO) $ & & & \checkmark & \checkmark & & \multicolumn{2}{c}{}\\
$x_4 (SO) \leftarrow x_3 (FE) $ & & & \checkmark & & \checkmark & \multicolumn{2}{c}{}\\
$x_4 (SO) \leftarrow x_5 (SE)$ & & \checkmark & & & \checkmark & \multicolumn{2}{c}{}\\
$x_4 (SO) \leftarrow x_6 (NS) $ & & \checkmark & & & & \multicolumn{2}{c}{}\\
$x_5 (SE) \leftarrow x_1 (FO)$ & \checkmark & & \checkmark & \checkmark & & \multicolumn{2}{c}{}\\
$x_5 (SE) \leftarrow x_3 (FE) $ & & & \checkmark & & \checkmark & \multicolumn{2}{c}{}\\
$x_5 (SE) \leftarrow x_6 (NS) $ & & & & & & \multicolumn{2}{c}{}\\
{\color{black} $x_6 (NS) \leftarrow x_1 (FO)$} & \checkmark & & \checkmark & & & \multicolumn{2}{c}{}\\
{\color{black} $x_6 (NS) \leftarrow x_3 (FE)$} & \checkmark & & \checkmark & & \checkmark &\multicolumn{2}{c}{} \\
\hline
Num. of successes & 3 & 8 & 9 & 2 & 9 & \multicolumn{2}{c}{}\\
Precisions & 0.20 & 0.53 & 0.60 & 0.13 & 0.60 & \multicolumn{2}{c}{}\\
\multicolumn{8}{l}{}\\
\multicolumn{2}{l}{FO: Father's Occupation} &
\multicolumn{6}{l}{ICA: ICA-LiNGAM \citep{Shimizu06JMLR}}\\
\multicolumn{2}{l}{FE: Father's Education} &
\multicolumn{6}{l}{Direct: DirectLiNGAM \citep{Shimizu11JMLR}}\\
\multicolumn{2}{l}{SI: Son's Income} &
\multicolumn{6}{l}{Pairwise: Pairwise~LiNGAM \citep{Hyva13JMLR}}\\
\multicolumn{2}{l}{SO: Son's Occupation} &
\multicolumn{6}{l}{PNL: Post-nonlinear causal model}\\
\multicolumn{2}{l}{SE: Son's Education} &
\multicolumn{6}{l}{\citep{Zhang09UAI}}\\
\multicolumn{2}{l}{NS: Number of Siblings} &
\multicolumn{6}{l}{}\\
\end{tabular}
\label{tab:Bollen}
\end{center}
\end{table}%

\begin{table}[]
\begin{center}
\caption{Estimated hyper-parameter values of our method {\color{black}
with $t$-distributed individual-specific effects}}
\scriptsize
\begin{tabular}{lc|crrr}
Pairs analyzed & Possible & Estimated & $\hat{\tau}^{indvdl}_1$ & $\hat{\tau}^{indvdl}_2$ & $\hat{\sigma}_{12}$\\
& directions & directions & & & \\
{\color{black} ($x_1 (FO), x_3 (FE)$)} & $\leftarrow$ & $\leftarrow$ & $0.4^2\widehat{\rm var}(x_1)$ & 0 & -0.7\\
($x_2 (SI), x_1 (FO)$) & $\leftarrow$ & $\leftarrow$ & $0.8^2\widehat{\rm var}(x_2)$ & 0 & 0.3 \\
($x_2 (SI), x_3 (FE) $) & $\leftarrow$ & $\leftarrow$ & $0.8^2\widehat{\rm var}(x_2)$ & 0 & -0.5 \\ 
($x_2 (SI), x_4 (SO)$) & $\leftarrow $ & $\leftarrow$ & $0.2^2\widehat{\rm var}(x_2)$ & $0.4^2\widehat{\rm var}(x_4)$ & -0.5 \\ 
($x_2 (SI), x_5 (SE)$) & $\leftarrow$ & $\leftarrow$ & 0 & $0.4^2\widehat{\rm var}(x_5)$ & 0 \\
($x_2 (SI), x_6 (NS)$) & $\leftarrow$ & $\leftarrow$ & $1.0^2\widehat{\rm var}(x_2)$ & $0.6^2\widehat{\rm var}(x_6)$ & 0 \\
($x_4 (SO), x_1 (FO)$) & $\leftarrow$ & $\leftarrow$ & $1.0^2\widehat{\rm var}(x_4)$ & 0 & 0.9 \\ 
($x_4 (SO), x_3 (FE)$) & $\leftarrow$ & $\leftarrow$ & 0 & $0.2^2\widehat{\rm var}(x_3)$ & -0.3 \\ 
($x_4 (SO), x_5 (SE)$) & $\leftarrow$ & $\leftarrow$ & 0 & 0 & -0.3 \\
($x_4 (SO), x_6 (NS)$) & $\leftarrow$ & $\leftarrow$ & $0.6^2\widehat{\rm var}(x_4)$ & 0 & -0.7 \\ 
($x_5 (SE), x_1 (FO)$) & $\leftarrow$ & $\rightarrow$ & 0 & $0.8^2\widehat{\rm var}(x_1)$ & 0.3 \\ 
($x_5 (SE), x_3 (FE)$) & $\leftarrow$ & $\leftarrow$ & $0.6^2\widehat{\rm var}(x_5)$ & 0 & -0.5 \\
($x_5 (SE), x_6 (NS)$) & $\leftarrow$ & $\leftarrow$ & $0.2^2\widehat{\rm var}(x_5)$ & 0 & -0.3 \\ 
{\color{black} ($x_6 (NS), x_1 (FO)$)} & $\leftarrow$ & $\rightarrow$ & $0.2^2\widehat{\rm var}(x_6)$ & 0 & -0.9\\
{\color{black} ($x_6 (NS), x_3 (FE)$)} & $\leftarrow$ & $\rightarrow$ & $0.2^2\widehat{\rm var}(x_6)$ & $0.6^2\widehat{\rm var}(x_3)$ & 0.5 \\
\multicolumn{6}{l}{}\\
\multicolumn{6}{l}{FO: Father's Occupation}\\ 
\multicolumn{6}{l}{FE: Father's Education}\\
\multicolumn{6}{l}{SI: Son's Income}\\
\multicolumn{6}{l}{SO: Son's Occupation}\\
\multicolumn{6}{l}{SE: Son's Education}\\
\multicolumn{6}{l}{NS: Number of Siblings}\\
\multicolumn{6}{l}{}\\
\multicolumn{6}{l}{$\tau^{indvdl}_1$ and $\tau^{indvdl}_2$ represent the variances of the individual-specific effects for}\\
\multicolumn{6}{l}{the variable pairs in the left-most column.}\\
\multicolumn{6}{l}{$\sigma_{12}$ represents the correlation parameter value of the individual-specific effects for}\\
\multicolumn{6}{l}{the variable pairs in the left-most column.}
\end{tabular}
\label{tab:esthyperparameters}
\end{center}
\end{table}%

\section{Conclusions}\label{sec:conc}
We proposed a new variant of LiNGAM that incorporated individual-specific effects in order to allow latent confounders.
We further proposed an empirical Bayesian approach to estimate the possible causal direction of two observed variables based on the new model. 
In experiments on artificial data and real-world sociology data, the performance of our method was better than or at least comparable to that of existing methods. 

{\color{black}
For more than two variables, one approach would be to apply our method on every pair of the variables. Then, we can estimate a causal ordering of all the variables by integrating the estimation results. 
{\color{black} This approach is computationally much simper than trying all the possible causal orderings.} 
Once a causal ordering of the variables is estimated, the remaining problem is to estimate regression coefficients or their posterior distributions. Then, one can see if there are direct causal connections between these variables. Although this could still be computationally challenging for large numbers of variables, the problem reduces to a significantly simpler one by identifying their causal orders. Thus, it is sensible to develop methods that can estimate causal direction of two variables allowing latent confounders. 
}

Future work will focus on extending the model to allow cyclic and nonlinear relations and a wider class of non-Gaussian distributions as well as evaluating our method on various real-world data.
Another important direction is to investigate the degree to which the model selection is sensitive to the choice of prior distributions.

\subsubsection*{Acknowledgments.}
S.S. was supported by KAKENHI \#24700275. 
We thank Aapo Hyv\"arinen and Ricardo Silva for their helpful comments. 

\bibliography{shimizu13a_bib}
\bibliographystyle{apalike}

\end{document}